\documentclass{article}

\PassOptionsToPackage{numbers, compress}{natbib}
\usepackage[final]{neurips_2018}

\usepackage{amsmath}
\usepackage{wrapfig}
\usepackage{amssymb}
\usepackage{shortbold}
\usepackage{lib}
\usepackage{lib2}
\usepackage{verbatim}
\usepackage{graphicx}
\usepackage[font=footnotesize,labelfont=bf,skip=2pt]{caption}
\usepackage{color}
\usepackage{multirow}
\usepackage[utf8]{inputenc} 
\usepackage[T1]{fontenc}    
\usepackage[pagebackref=true,breaklinks=true,letterpaper=true,colorlinks,bookmarks=false,citecolor=blue]{hyperref}
\usepackage{url}            
\usepackage{booktabs}       
\usepackage{amsfonts}       
\usepackage{nicefrac}       
\usepackage{microtype}      

\title{Visual Memory for Robust Path Following} 

\author{Ashish Kumar\thanks{Equal contribution. Project website with videos: \url{https://ashishkumar1993.github.io/rpf/}.}
\hfill Saurabh Gupta$^*$ \hfill David Fouhey \hfill Sergey
Levine \hfill Jitendra Malik \\
University of California, Berkeley \\
\texttt{\small ashish\_kumar@berkeley.edu, \{sgupta, dfouhey, svlevine, malik\}@eecs.berkeley.edu}}

\begin{document}

\newcommand{\note}[1]{}

\maketitle

\begin{abstract}
Humans routinely retrace paths in a novel environment both forwards and
backwards despite uncertainty in their motion. This paper presents an
approach for doing so. Given a demonstration of a path, a first network
generates a path abstraction. Equipped with this abstraction, a second
network observes the world and decides how to act to retrace the
path under noisy actuation and a changing environment. The two networks are
optimized end-to-end at training time. We evaluate the method in two realistic
simulators, performing path following and homing under actuation noise and
environmental changes. Our experiments show that our approach outperforms
classical approaches and other learning based baselines.
\end{abstract}
\section{Introduction}
Consider the first morning of a conference in a city you have never been to.
Rushing to the first talk, you might follow your phone's directions through a
series of twists and turns to reach the venue. When you return later in the
day, you can retrace your steps to your hotel relatively robustly, remembering
to take a left turn at the bistro and keep straight past the coffee shop. The
next day, you may probably only look at your phone to check your email. At
first glance, this seems like a trivial ability. Humans routinely do this, for
instance when a friend shows you the bathroom in their apartment or when you go
to a room in a new building. On second glance though, it is an amazing ability 
since one never retraces one's steps exactly and the visual experience
is constantly changing in fairly dramatic ways: people move their cars, shops
open and close, and seasons change. This paper aims to replicate
this ability to retrace paths (including reversals) in new environments
with imperfect ability to replicate one's actions (actuation) as well as
a changing world.

How might we solve this problem? One classical approach, common in robotics,
would be to build a full 3D model of the world via SLAM, from building facades
to the side-mirrors of cars, during the first pass; after this is done, path
following reduces to localizing in the model and selecting the best action. For
the task of navigation, this is simultaneously too much work -- precisely
reconstructing the facade is less important than recognizing it as ``the bistro
at which I turn left'' -- as well as too little work -- the parts of the
reconstruction that provide stable localization and the parts that do not are
mixed together with no way to disentangle them.

Given these difficulties of classical approaches, a large number of
learning-based approaches have sprung up to solve this and other related
navigation tasks (\eg \cite{mirowski2016learning, mirowski2018learning,
gupta2017cognitive, savinov2018semi, zhu2016target, parisotto2017neural}). In
these works, an agent directly predicts actions from image observations in an
end-to-end fashion. However, unlike our setup of a single demonstration in a
new environment, many of these setups require the agent to have a great deal of
experience with the \textit{test} environment via reward based supervision
\cite{mirowski2018learning, mirowski2016learning, zhu2016target} or exhaustive
human demonstrations \cite{savinov2018semi}.  Moreover, unlike most work on
navigation in new environments \cite{gupta2017cognitive}, our setup poses the
additional challenge of noisy actuation as well as a world that can change
between the initial demonstration and path execution.

Our approach, which we describe in \secref{approach}, consists of a module that
learns to convert a series of observations of a path to an abstract
representation, and a learned controller that \textit{implicitly} localizes the
agent along this abstracted path using the current observation and outputs
actions that bring the agent to the desired goal location. We see a number of
advantages to training the whole approach end-to-end on data in comparison to
classical approaches.  First, the learned model can use statistical
regularities to make its performance more robust: for example, it can learn to
count doors in a texture-less hallway rather than localize at each point and
when it returns along a path, it can learn to look on the right for a table
that was previously on the left. Second, by virtue of being learned entirely
end-to-end, rather than being learned and designed piecemeal, the approach can
automatically learn the features that are necessary for the task at hand. As a
concrete demonstration, we evaluate a homing task in which the agent retraces a
path in reverse; the network learns to produce features necessary for solving
this task without explicitly designing any wide-baseline features or proxy
tasks.

We evaluate our approach on multiple datasets in \secref{exp} in a series of
experiments that aim to probe to what extent we can learn to retrace a path
under noisy actuation and in a changing world. We compare to a variety of
classical and learned alternate approaches, and outperform them. In particular,
our experimental results show the value of end-to-end learning the entire
path-following process.

\begin{figure}
  \centering
  \includegraphics[width=\textwidth]{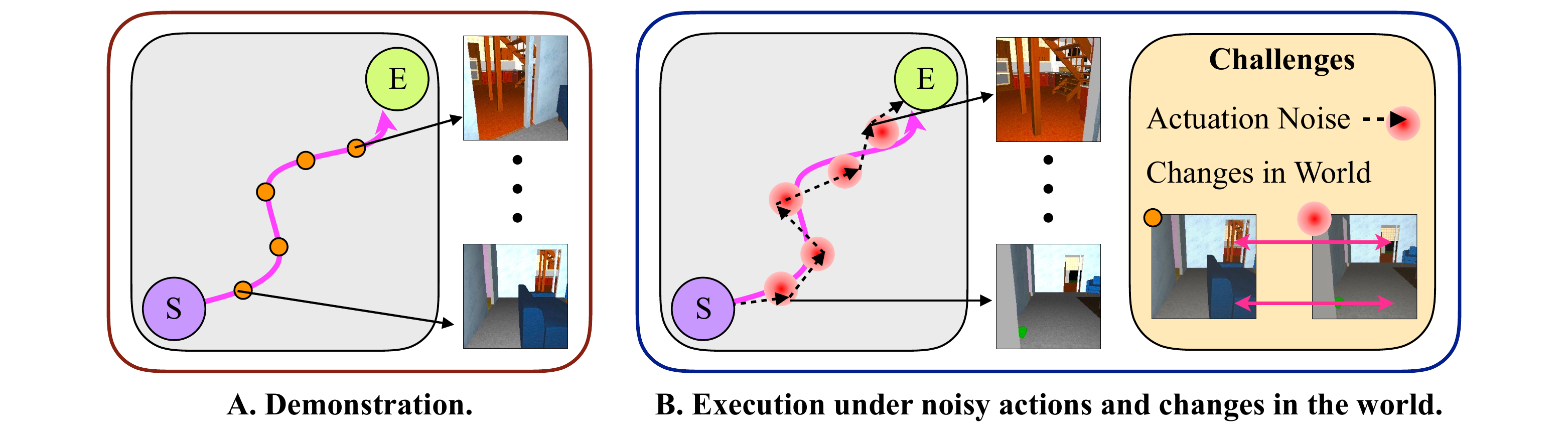}
  \caption{\textbf{Problem Setup:} Given images and associated actions from a
  traversed path (A), we want to retrace the path both forward and backward
  under actuation noise (\ie uncertainty in movement) in a changing world (\eg
  the chairs have been moved between the demonstration and execution) (B). This
  paper presents an end-to-end neural architecture for this task. We use the
  demonstration path to generate a path abstraction that is used by a learned
  controller to reliably convey the agent to the desired target location at
  execution time.}
  \figlabel{overview}
  \vspace{-16pt}
\end{figure}

\section{Related Work}
\seclabel{related}

In this work, we study the problem of retracing a path under noisy actuation
and a changing world. This touches on classical work in robotics and recent
works learning-based efforts. 

One classic approach to solving the path retracing problem (also referred to as
Visual Teach and Repeat \cite{furgale2010visual}) is to chain together core
robotics primitives of mapping, localizing, and planning: on the initial
demonstration, the agent builds a 3D map; during subsequent navigation
attempts, the agent localizes itself in the map and generates actions
accordingly. Each problem has been studied extensively in robotics, typically
focusing on geometric solutions, \ie  using metric maps and locations. For
example, mapping has been often studied as the classic simultaneous
localization and mapping (SLAM) problem
\cite{davison1998mobile,engel2014lsd,slam-survey:2015,izadiUIST11} in which one
builds a 3D map of the world. Localization is often framed as recovering camera
pose with respect to a global or local 3D map \cite{clement2017robust,
cummins2008fab} or by performing visual odometry \cite{nister2004visual,
zhou2017unsupervised} to obtain a metric displacement from a start position.
Finally, planning is often done assuming direct access to a noiseless map and
often a noiseless agent location
\cite{canny1988complexity,kavraki1996probabilistic,lavalle2000rapidly}.

The distinction from this line of work is that our approach is learned
end-to-end. This means that intermediate representations are automatically
learned, rich non-geometric information can be incorporated, and modules are
optimized jointly for the end-goal. Obviously, researchers are aware of the limitations of
purely geometric and pipelined approaches, and have developed extensions:
explicitly incorporating semantics into SLAM \cite{bowman2017probabilistic},
adding margins in planning to account for uncertainty
\cite{axelrod2017provably}, and using learning for sub-sets of the SLAM problem
\cite{brahmbhatt2017mapnet, chaplot2018active, kar2017learning,
kendall2015posenet, pillai2017towards, schonberger2017semantic,
swedish2018deep}. The strength of our proposed method is that these strategies
are learned automatically and are specified by what is needed empirically to
navigate an environment as opposed to human intuition.

We are not the first to recognize the potential for end-to-end task-driven
navigation. In recent years there has been a flurry of work in this area. A lot
of this work focuses on the shorter time-scale task of collision avoidance, \eg
how do I move through the door without bumping into it
\cite{daftry2016learning, gandhi2017learning, giusti2016machine, kahn2017self,
pomerleau1989alvinn, sadeghi2016cadrl}.  In contrast, our work focuses on the
longer time-scale problem of path following \eg how do I get back to my office
from the coffee machine. At this time-scale, some works have framed the problem
of navigation as learning to reach different goals in a fixed training
environment \cite{mirowski2018learning, mirowski2016learning}, designing
policies that can directly act in new test environments
\cite{gupta2017cognitive, khan2017memory, parisotto2017neural,
zhang2017neural}, self-supervised learning for reaching goals in a
well-traversed environment \cite{savinov2018semi} or by following a
demonstration \cite{pathak2018zeroshot}.  Our work is more similar to this last
line of work \cite{pathak2018zeroshot, savinov2018semi}.  It is distinguished,
however, by its lack of explicit localization as well as its navigation under
{\it actuation noise} and {\it changing environments}. This noise is a crucial
distinction because noiseless execution of actions and an unchanged environment
means that memorization alone is sufficient. We also study the task of homing, where
no direct demonstration images are available.

\begin{figure}
    \centering
    \includegraphics[width=\textwidth]{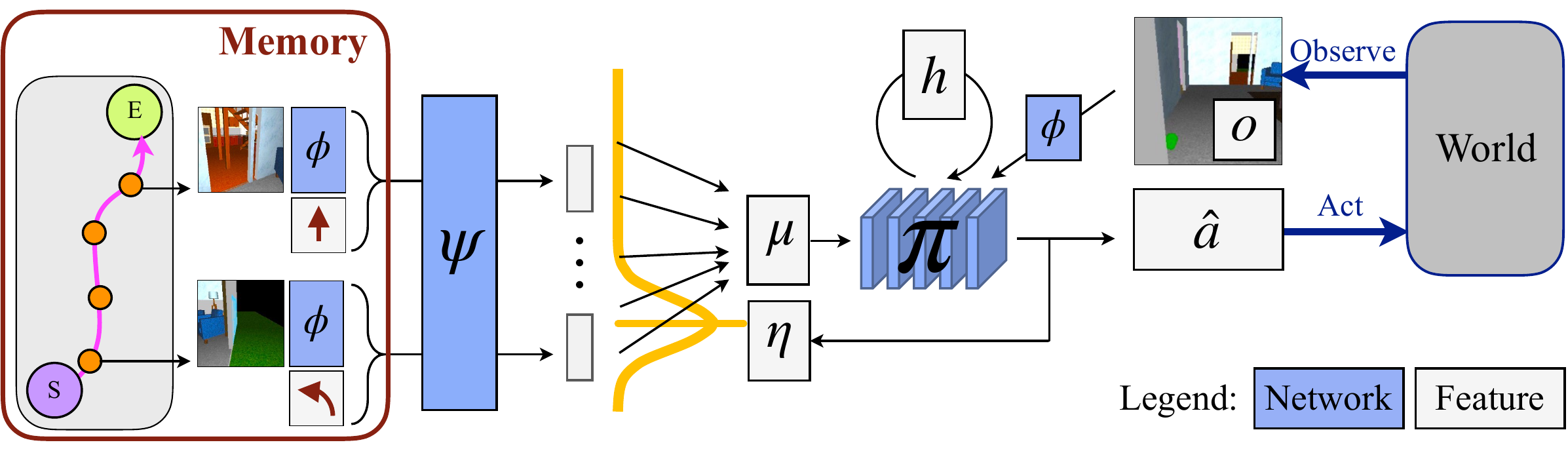} \caption{{\bf
    Proposed Approach:} As input, our model takes a sequence of images
    (processed by $\phi$) and actions at these images $\aB(\pB)_j$. It
    abstracts this sequence into a sequence of memories. A second, recurrent,
    network $\pi$ uses this sequence of memories to emit actions that retrace
    the path. At each time-step, $\pi$ reads in: (a) the sequence, softly
    attending to the relevant part at past time-step via $\eta$; (b) the current
    observation $O$ from the world (also processed by $\phi$); as well as its
    recurrent hidden state $h$. As output, $\pi$ updates the attention
    location $\eta$ and its hidden state, and emits action $\hat{a}$ that the
    agent executes in the world.}
    \figlabel{executing}
    \vspace{-15pt}
\end{figure}

\section{Robust Path Following}
\seclabel{approach}

\textbf{Problem Setup.}
Consider an agent that we are trying to train to operate in a new
environment $E$ that it has never been in before. 
Let us assume that the agent's state at time $t$ is represented by $s_t$, and
that the agent has some primitive set of \textit{stochastic} actions
$\mathcal{A}$ that it can execute. Executing an action $a \in \mathcal{A}$,
takes the agent to state $s_{t+1}$ via a stochastic transition function $f$ \ie
$s_{t+1}$ is sampled randomly from the distribution $f(s_{t}, a, E)$. We assume
that the agent is equipped with a first person \rgb camera to obtain visual
observations of the environment $I = \rho(E, s)$, where the function
$\rho$ renders an environment $E$ from agent's current location specified by $s$.
We do not tackle the problem of higher-level planning and abstract away 
low-level motor control to focus on the problem of robust path following.
Our approach can be used with classical (or even learned) approaches for
path planning and low-level motor control.

Suppose we move this agent around in this new environment along a path
$\pB$. Given such a traversal, we want the agent to be able to reliably
re-trace the path $\pB$ or to follow a related path such as the reversed
path (denoted $\tilde{\pB}$). We want to be able to do this in situations where the agent has
noisy actuation and sensing, and the environment changes (to $E'$) between the
demonstration of $\pB$ and when the agent is tested to autonomously traverse
$\pB$ (or $\tilde{\pB}$). Note that, the only sensory information that is available to
the agent as it is trying to traverse the path $\pB$ (or $\tilde{\pB}$) in the
environment $E$ (or $E'$) is via $\rho$. It does not have access to the ground
truth state of the system $s_{t}$, but only \rgb image observations of the
environment from that state.

Our goal is to learn a policy $\Pi$ that will achieve this. As input, $\Pi$
assumes access to the: path $\pB$; actions $\aB(\pB)$ taken while executing
the path; visual observations along the path 
$\IB = \{I_1 \ldots I_J : I_j = \rho(E, \pB_j)\}$;
as well as the current visual observation $O$. As output, $\Pi$ predicts
actions from $\mathcal{A}$ that successfully and
efficiently convey the agent to the destination $\pB_J$, \ie $\hat{a} =
{\Pi(\pB, \IB, O)}$. This action $\hat{a}$ is executed in the stochastic
environment; the agent obtains the new visual observation at the next state;
and this process is repeated for a fixed number of time steps.

We now describe the policy function $\Pi$. Intuitively, our policy function
$\Pi$ uses the observation $O$ to implicitly localize the agent with
respect to its memory of the path. Given this localization, the policy 
reads out relevant information (such as relative pose and actions) and uses
this in context of $O$ to take an action that conveys the agent
to the desired target location. The entire policy is implemented using
neural network modules that are differentiable and learned end-to-end
using training data for the task of efficiently going to
the desired goal location.

We first describe the basic architecture for $\Pi$ where we want to retrace the
same path $\pB$ that was demonstrated and then describe the extension to
retrace a related path $\tilde{\pB}$. We call our proposed policy $\Pi$ as
Robust Path Following policy and denote it by \rpf. 

\subsection{Learned Controller for Robust Path Following}
\seclabel{exec}
The policy $\Pi(\pB,\IB,O)$ is realized as follows. We first use $\pB$ and
$\IB$ to compute a path description $M(\pB)$ that captures the local
information needed to follow the path. $M(\pB)$ is a sequence of tuples
consisting of features of the reference image, associated reference position
and the associated reference action for each step $j$ in the trajectory, or
\begin{eqnarray}
\eqlabel{M}
M(\pB) &=& \{\left(\aB(\pB)_j, {\phi(I_j)}\right) : j \in [1 \ldots J]\} .
\end{eqnarray}
We use this path description with a learned controller that takes as input the
current image observation $O$ to output actions to follow the path under noisy actuation.
We represent the policy $\pi$ with a recurrent neural network that iterates over
the path description $M(\pB)$ as the agent moves through the environment. This
iteration is implemented using attention that traverses
over the path description. At each step, the path signature is read into $\mu_t$ with
differentiable soft attention centered at $\eta_t$:
\begin{eqnarray}
\vspace{-0.05in}
\mu_t &=& \sum_j \psi\left(M(\pB)_j\right) e^{-|\eta_t-j|}. 
\vspace{-0.05in}
\end{eqnarray}
The recurrent function $\pi$ with state $h_t$ is implemented as:
\begin{eqnarray}
\vspace{-0.05in}
h_{t+1}, b, \hat{a} = \pi(h_t, \mu_t, \phi(O)) \mbox{ and } \eta_{t+1} = \eta_{t} + \sigma(b).
\vspace{-0.05in}
\end{eqnarray}
As input, it takes the internal state, $h_t$, attended path signature $\mu_t$
and featurized image observation $\phi(O)$. In return, it gives a new state
$h_{t+1}$, pointer increment $b$, and action $\hat{a}$ that the agent should
execute. This pointer increment is passed through an increment function
$\sigma$, and added to $\eta_{t}$ to yield the new pointer $\eta_{t+1}$.
Given the agent moves by one step per action in expectation, we use
$1+\text{tanh}$ as the increment function $\sigma$. We set $\eta_1 = 1$ and $h_1 = \mathbf{0}$.

Note that we factor out the controller from the path description. This
factorization of the environment and goal specific information into a path
description that is separate from the policy lets us learn a \textit{single}
policy $\pi$ that can do different things in different environments with
different path descriptions without requiring any re-training or adaptation. The
policy can then also be thought of as a robust parameterized goal-oriented
closed-loop controller.

\subsection{Feature Synthesis for Following Related Paths}
\seclabel{synth}
So far our approach can only repeat the paths that we have
already taken, but our approach can be extended to follow paths $\tilde{\pB}$ that are
related to but not the same as the path $\pB$ that was demonstrated.
We do this by {\it synthesizing} features for the path $\tilde{\pB}$ using whatever
information is available for the demonstrated path $\pB$.
We synthesize features $\hat{\phi}(\tilde{\pB}_j)$ for location $\tilde{\pB}_j$ using
observed features $\phi(I_i) $ from location $\pB_i$ as follows:
\begin{eqnarray}
\omega_{i,j} = \Omega\left(\left(\phi(I_i), \delta(\pB_i, \tilde{\pB}_j)\right)\right) \mbox{ and } \hat{\phi}(\tilde{\pB}_j) = \Sigma_g \left(\omega_{1,j}, \omega_{2,j}, \ldots, \omega_{N,j}\right) . 
\end{eqnarray}
Here, the function $\delta$ computes the relative pose of image $I_i$ with
respect to the desired synthesis location $\tilde{\pB}_j$. $\phi$ computes the
representation for image $I_i$ through a CNN followed by two fully connected
layers. $\Omega$ fuses the relative pose with the representation for the image
to obtain the contribution $\omega_{i,j}$ of image $I_i$ towards representation at
location $\tilde{\pB}_j$. These contributions $\omega_{i,j}$ from different images are
accumulated through a weighted addition by function $\Sigma_g$ to obtain the
synthesized feature $\hat{\phi}(\tilde{\pB}_j)$ at location $\tilde{\pB}_j$. The path
description $M(\tilde{\pB})$ can then be obtained as a collection of tuples
$(\aB(\tilde{\pB})_j, \hat{\phi}(\tilde{\pB}_j))$.

{\bf Implementation Details.} We now describe the particular architecture that
we use throughout the paper.  $\phi$ is a 5 layer Convolutional Network with
[32, 64, 128, 256, 512] filters respectively. Each conv layer is followed by a
max-pooling.  $\psi$ and $\Omega$ are fully connected networks consisting of two
layers; $\pi$ is implemented using GRUs. We train the entire network from
scratch in an end-to-end manner using Adam optimizer for 120000 iterations,
where each episode is 40 steps long. 

\section{Experiments}
\seclabel{exp}
This paper studies the task of retracing a route in a {\it new} environment
(either forwards or backwards) under noisy actuation and a changing world,
given a demonstration. Our experiments evaluate: a) to what extent can we solve
this task, b) what is the role of visual memories, and c) how our proposed
solution compares to classical geometry-based and other learning-based
solutions. Crucially, we further study how our policies perform on
settings outside of what they were explicitly trained on.

\subsection{Experimental setup}
\noindent {\bf Simulators.} We use two simulation environments that permit
rendering from arbitrary viewpoints and allow separation of held-out
environments for testing.
The first simulator is based on real world scans from {\it Stanford Building
Parser Dataset} \cite{armeni20163d} (\sbpd) and the {\it Matterport 3D Dataset}
\cite{Matterport3D}(\mpd). These scans have been used to study navigation tasks
in \cite{gupta2017cognitive}, and we adapt their publicly available simulation code.
We use splits that ensure that the testing environment comes from an entirely
different building than the training or validation environments. 
The second simulation environment is based on {\it SUNCG} \cite{song2016ssc}.
SUNCG consists of synthetic indoor environments with manually created room and
furniture layouts that have corresponding textured meshes \cite{song2016ssc}.
Because these environments are graphics codes, SUNCG permits evaluation of the
effect of environmental changes. In particular, objects can be removed without
inducing artifacts that a network will pick up on. Once again, splits ensure no
overlap between training and testing environments.

\noindent {\bf Agent Actions.} We assume that the agent has 4 macro-actions,
stay in place, rotate left or right by $\theta$ ($=30^\circ$), and move forward
$x$ units ($=40cm$). 

\noindent {\bf Noise Model.} Our work studies path retracing both with
actuation noise (\ie the outcome of actions is stochastic) and a changing world
(the world changes between the demonstration and autonomous operation of the
agent).  {\it Actuation Noise:} In both environments, when the agent outputs
the rotation actions it actually rotates by ${\sim}N_{trunc}(\theta,
57.3^\circ, 0.2|\theta|)$. Here, $ N_{trunc}(\mu, \sigma, \delta)$ refers to a
normal distribution with mean $\mu$ and standard deviation $\sigma$ that is
truncated to $(\mu-\delta, \mu+\delta)$. When the agent executes a move forward
action it rotates by ${\sim}N_{trunc}(0, 57.3^\circ, 0.2|\theta|)$ and then
translates by ${\sim}N_{trunc}(x, 5cm, 0.2x)$.  The factor of $0.2$ controls
the amount of noise in actuation and we vary it in our experiments.  {\it World
Changes:} World changes are studied in the SUNCG environment.  Demonstrations
are provided in an environment with objects removed uniformly with a
probability of $r$ ($=.5$). The task is to get to the desired target location
in presence of even fewer objects ($r$ is $.1$, or $.3$) or additional objects
($r$ is $.7$, or $.9$).

\noindent {\bf Tasks.} Given a path $\pB$ from $\pB_0$ to $\pB_T$, we consider
the two tasks of {\it trajectory following} \ie going from $\pB_0$ to $\pB_T$
as well as {\it homing} \ie going from $\pB_T$ to $\pB_0$. We evaluate these
tasks under a variety of noise conditions and environments.

\noindent {\bf Evaluation Criteria.} We characterize the success of the agent
by measuring how close it gets to the goal location. We analyze each approach
over $500$ trials in a novel environment (not used for training).  We report
three metrics: a) {\it Success Rate} (reaching within 2 steps or $10\%$ of the
initial distance to goal, whichever is larger), b) {\it \spl} (Success weighted
by normalized inverse Path Length as described in
\cite{anderson2018evaluation}), and c) {\it Median Normalized Distance} to goal
at end of episode.

\noindent {\bf Model Training.} We use imitation learning to train our
policies. Although the agent never has access to its true location as it
traverses the environment, the true location is available in the
simulator. This is used to compute a set of `good actions' that will convey the
agent to the desired target location. This set of good actions are ones that
lead to a larger reduction in the distance to goal when compared to forward
action. We optimize the policy to minimize the negative log of the sum of
probabilities of all good actions at each time step.

\subsection{Baselines}
We compare with a number of baselines that represent either classical
approaches or test the importance of various components of our system.

{\bf Open Loop.} We repeat the reference actions (or their reverse for homing).
Under perfect actuation this would achieve perfect performance. With actuation
noise, this baseline is a measure of the hardness of the task and tests to what
extent learning to act under actuation noise is necessary. 

{\bf Visual Servoing.} For each action, we compute the $L2$ distance between
SIFT feature matches of target reference image (initially set to first
reference image) and image expected after executing that action (we obtain this
image by {\it virtually} executing this step in the simulator). The policy
actually executes the action that has the lowest distance. To decide when to
increment the target image, we check the ground truth proximity to the next
reference image. We stop when the agent reaches the last target image.  

{\bf 3D Reconstruction and Localization.} We use the
publicly available COLMAP package \cite{schoenberger2016sfm,
schoenberger2016mvs} that implements a variety of geometric mapping and
localization algorithms. Note that these geometry-based methods require
high-resolution images at high frame rates. Thus, we sample high-resolution
images ($1024\times 1024$ \vs $224 \times 224$ for our policies) at $5\times$
the frame rate ($145$ \vs $30$ images for our policy for a trajectory of length
$30$) along the reference trajectory along with ground truth poses. This
ensures that reconstruction via SIFT key-point matching
\cite{lowe2004distinctive} and bundle adjustment always succeeds.
Given this reconstruction, the agent localizes itself by registering SIFT
key-points on the current image with the 3D reconstruction derived from 
the reference images. It then estimates free space by marking a small 
region around each point on the reference trajectory as free. Given this inferred
free space and localization, it executes the action that most efficiently 
conveys it to the goal location. If the localization fails, it rotates 
left at each step until localization succeeds (or the episode ends). 

{\bf \rpf with No Visual Memory.} We also compare to a policy without any
visual memory. We do this by only using the action and pose as part of the
memory $M(\pB)$ in \eqref{M} \ie removing $\phi(I_j)$. We retain the rest of
the architecture of \rpf. Note that this is a competitive comparison as
the policy can learn to replay actions meaningfully in context of current
images from the environment.

All learned policies are trained in \textit{base settings}: the noise level is
at $20\%$, reference trajectories are of length $30$, the policy is executed
for $40$ time steps, and these reference trajectories are sampled to be far
from obstacles. \figref{results} and \tableref{base-results} present our
experimental results. We report the success rate, \spl, and median normalized
distance and compare against the baselines described above. In addition, we
also show a bootstrapped $95\%$ confidence interval for each plot. 

\renewcommand{\arraystretch}{1.1} 
\setlength{\tabcolsep}{6pt}
\begin{table}
\centering
\footnotesize
\caption{Performance over $500$ trials on the test set (\sbpd \texttt{area4})
in base settings for the Following and Homing tasks. We report Success Rate,
\spl and Median Normalized Distance. See text for details.}
\tablelabel{base-results}
\resizebox{0.9\linewidth}{!}{
\centering
\begin{tabular}{lcccccc}
\toprule
& Open Loop & Visual Servoing & 3D Recons. + Localize & RPF (no visual mem.) & RPF
\\ \midrule
\multicolumn{2}{l}{\textbf{Following}} 
\\ $\quad$ Success Rate $\uparrow$                 & 0.216 & 0.318 & 0.826 & 0.782 & \textbf{0.866} 
\\ $\quad$ SPL $\uparrow$                          & 0.191 & 0.293 & \textbf{0.766} & 0.648 & 0.726 
\\ $\quad$ Median Normalized Distance $\downarrow$ & 0.257 & 0.204 & 0.086 & 0.068 & \textbf{0.056} 
\\ \midrule
\multicolumn{2}{l}{\textbf{Homing}} 
\\ $\quad$ Success Rate $\uparrow$                 & 0.216 & -- & 0.000 & 0.780 & \textbf{0.866} 
\\ $\quad$ SPL $\uparrow$                          & 0.191 & -- & 0.000 & 0.644 & \textbf{0.740} 
\\ $\quad$ Median Normalized Distance $\downarrow$ & 0.257 & -- & 0.853 & 0.062 & \textbf{0.056} 
\\ \bottomrule    
\end{tabular}}
\vspace{-20pt}
\end{table}

\subsection{Results}

\textbf{Experiments on Matterport Data.} We first study the following and homing
tasks in static environments. As the environment does not need to change, we do
these experiments on realistic Matterport data. In particular, we train
policies on 4 floors from \sbpd and 6 buildings from \mpd. All policies are
tested on $\texttt{area4}$ from \sbpd which is from an altogether different
building than the 4 floors used for training. We found adding visually diverse
data from the \mpd dataset was crucial for good performance as our models are
trained entirely from scratch.

\textbf{Following Task.} \tableref{base-results}~(top) presents results for the
trajectory following task. 
Open loop replay only succeeds 22\% of the time. This is because
actuation noise compounds over time and causes both drift and collisions.
Visual feedback through servoing improves success rate to 32\%, though it is
limited as it only localizes against a single image at a time. This is
addressed by `3D Reconstruction + Localization' which uses all reference images
for localization (by reconstructing the environment), and achieves a success
rate of 83\%. \rpf achieves a higher success rate of 87\%. 
\rpf without visual memories succeeds 78\% of the time. This is a striking
result. Just the sequence of actions, when replayed intelligently through a
learned controller, leads to compelling path tracing performance. That is,
sequences of actions can robustly be interpreted in context of current image
observations without any explicit 3D reconstruction or localization.  Of
course, additionally using visual memories from the demonstrated path improves
performance.

\begin{figure}
\centering
\insertWL{1.0}{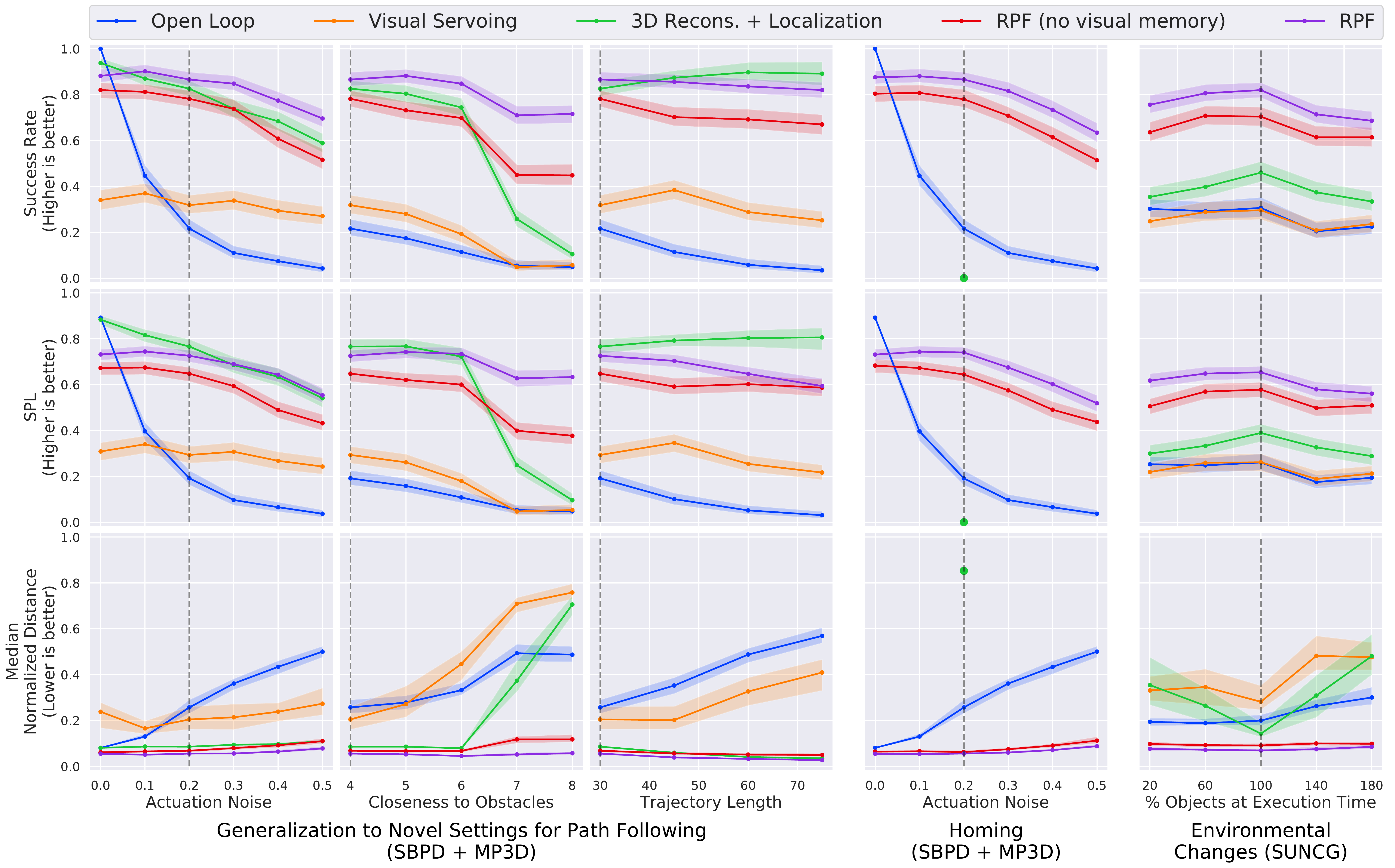}
\caption{{\bf Generalization Performance:} Rows show different metrics, and
columns show different test settings. Vertical dotted lines mark `base
settings' that the policies were trained on. 
We show generalization of \rpf to (far left) actuation noise levels, (center
left) obstacle distance, and (center) path length for path following, and
(center right) actuation noise for homing. The far right plot shows performance
under environmental changes.}
\figlabel{results} \vspace{-0.2in} \end{figure}

\textbf{Homing Task.} \tableref{base-results}~(bottom) shows results for the
homing task on Matterport data.  We set up the homing task by sampling the same
trajectories but simply providing images from a $180^\circ$ rotated camera at
each reference location. Thus open loop replay and \rpf without visual memories
perform about the same as before. Note that this scenario is impossible for
visual servoing as there are no direct images to servo to. While, in principle
3D Reconstruction and Localization can tackle this scenario, it performs poorly
($0\%$ success, $0.85$ median normalized distance) as SIFT features don't match
well across large baselines. In comparison, \rpf that learns to speculate
features still performs equally well at $87\%$. Note that this is better than
\rpf without visual memories, demonstrating that our feature prediction
technique is able to extract meaningful signal from related images.

\textbf{Testing on Out-of Train Settings.} We have shown so far that our
trained policies outperform appropriate baselines for the task when tested on
novel environments. However, we are still training and testing on the same
settings, such as noise level, trajectory length and distribution of
trajectories. We next test how well our learned policies work when we test them
on settings they have not been trained on. 
We do this by testing the policies trained in the `Base Setting', in different
settings in novel environments. We explore three novel settings: a)
\textit{Actuation Noise}: we vary the noise level of the environment, b)
\textit{Harder Trajectories}: we pick trajectories that are closer to obstacles
and require careful maneuverability, and c) \textit{Path Length}: we increase
the length of the trajectories to follow. We do not retrain our policies for
these settings, and simply execute the policy learned in the `Base Setting' on
these additional settings.

\figref{results}~(left 3 columns) presents the results. Rows plots the
different metrics (success rate, \spl, and median normalized distance), and
columns plot different test conditions. Plots in the first column show
performance as a function of actuation noise.  Policies trained at 20\% noise
are tested under noise varying between 0\% and 50\%. Open loop performs
perfectly with no noise, but its performance rapidly degrades as noise
increases. In comparison, \rpf performance degrades gracefully. Plots in the
second column show performance as we move to harder trajectories that are
sampled to be closer to obstacles (as we move from the left to the right on the
plot). \rpf degrades much more gracefully than the 3D reconstruction
based method. The third column shows performance as a function of trajectory
length. \rpf policies generalize reasonably well to even $3\times$ longer
trajectories than were seen during training. Column 4 shows similar plots of
performance variation as a function of the actuation noise for the homing task.
Finally, \figref{vis} shows multiple rollouts from our \rpf policy and
contrasts them with purely open loop rollouts.

\begin{figure}
\footnotesize
\centering
\setlength{\tabcolsep}{1pt}
\begin{tabular}{cccccc}
\multicolumn{2}{c}{{Following (Matterport)}} &
\multicolumn{2}{c}{{Homing (Matterport)}} & 
\multicolumn{2}{c}{{Changing Env (SUNCG)}} \\
\insertW{0.16}{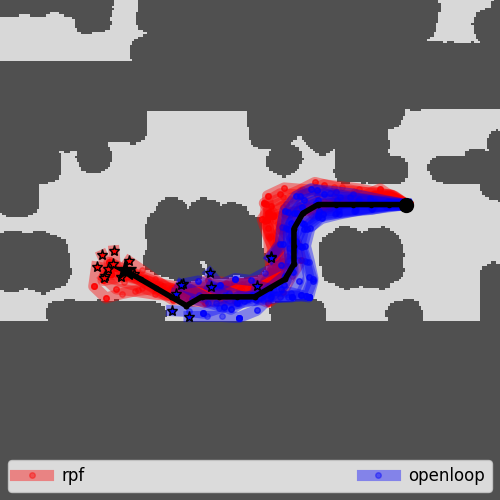} & 
\insertW{0.16}{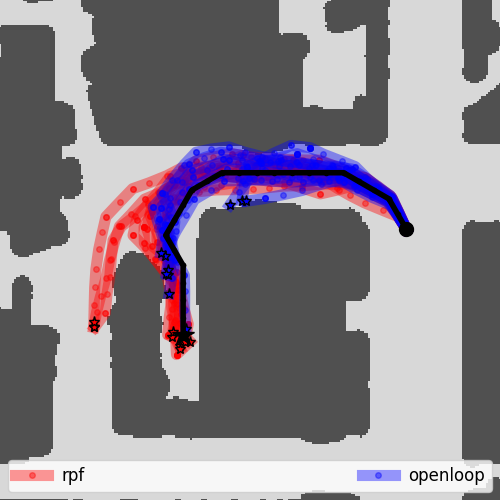} &
\insertW{0.16}{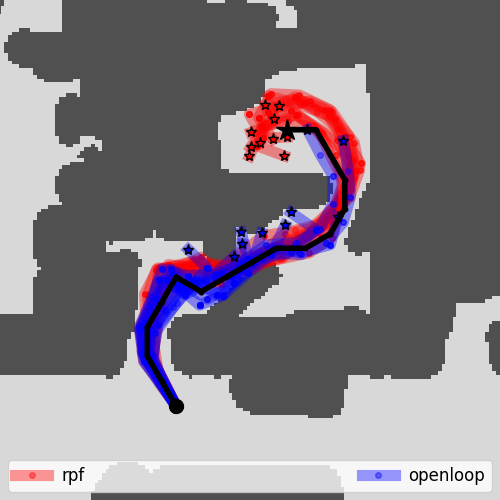} &
\insertW{0.16}{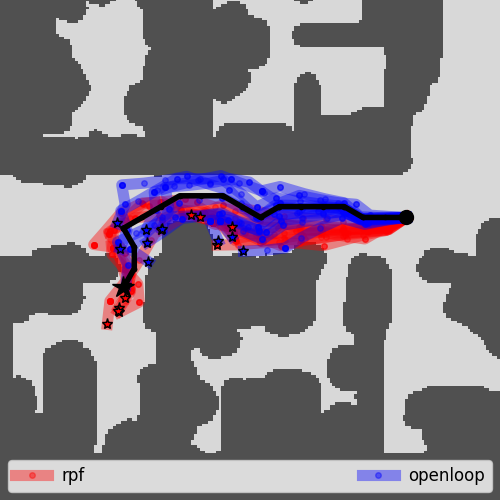} &
\insertW{0.16}{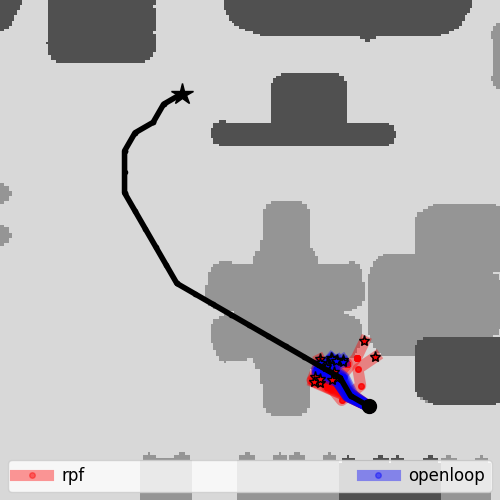} &
\insertW{0.16}{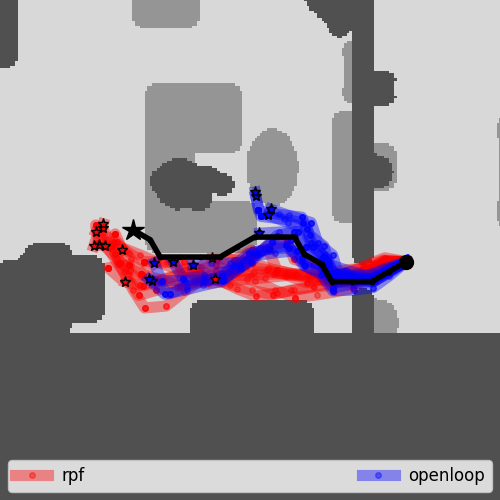}
\vspace{-2pt}
\end{tabular}
\caption{{\bf Trajectory Visualizations:} We visualize sample success and
failure trajectory in top view for the three different scenarios.
We show multiple roll-outs by resampling multiple times from the noise model.
Note that these top views are shown here only for visualization, the policy 
does not receive them and operates purely based on the first person views.
\rpf trajectories are in red, and open loop ones are in blue. 
\textbf{Following and Homing}: \rpf trajectories more tightly follow the
reference trajectory, while open loop roll-outs spread out and collide with
nearby obstacles. Overall, \rpf gets to the goal more reliably, but fails when
it drifts too far from the reference trajectory.  \textbf{Execution in Changing
Envirnonment in SUNCG:} Our approach is able to go around obstacles, but fails
if a very large deviation is required.}
\figlabel{vis}
\vspace{-20pt}
\end{figure}

\textbf{Experiments on SUNCG.} We next report experiments on SUNCG for the base
setting. We trained on 48 houses from the House3D training set and report
performance on 12 entirely disjoint houses from the test set.  We plot the
metrics in \figref{results}~(last column). The base setting (shown with the dotted
vertical line) here corresponds to when there are 100\% objects at execution
time \ie the environment is the same between when the reference trajectory was
provided, and when the agent has to execute the trajectory. As \rpf learns
features, it can better adapt to change in visual imagery in the synthetic
dataset as compared to feature-based geometric methods. Once again, \rpf was
trained in base settings.

\textbf{Robustness to Environmental Changes.} \figref{results}~(last column)
also plots performance as the environment changes. Points to the left of the
dotted line correspond to the setting where objects are removed from the
environment, while points to the right correspond to when objects are added
into the environment. In both these regimes, \rpf continues to perform well. In
contrast, performance for the 3D reconstruction and localization method degrades
sharply (see median normalized distance plot) as the environment at execution
time deviates from one at demonstration time. This is expected and known of
geometry based methods that do not cope well with changes in the environment. 
Moreover, the agent doesn't just need to be robust to visual changes between
reference images and the current observations, but it must also exhibit local
going-around-behavior as the reference trajectory may go through the newly
added obstacles. While \rpf is able to do so when there is a minor detour, it
fails when a much larger detour is required as shown in \figref{vis}~(right).

\textbf{Ablations and Comparisons to Other Learning Methods.} \figref{other}
reports success rate for the path retracing task for other learning methods:
a) \textit{Trained Nearest Neighbors}: we use similarity between current
observation with reference images to vote for the action, and b) \textit{GRU}:
a standard recurrent model without the proposed sequential memory. We also
compare to ablations of RPF: a) \textit{RPF without Recurrence}: $\pi$ is
implemented with a feed-forward network as opposed to a recurrent function, and
b) \textit{\rpf with a Constant Increment} instead of the learned increment
\setlength{\intextsep}{0pt}
\setlength{\columnsep}{8pt}
\begin{wrapfigure}[13]{r}{0.70\textwidth}
    \begin{center}
      \insertWL{1.0}{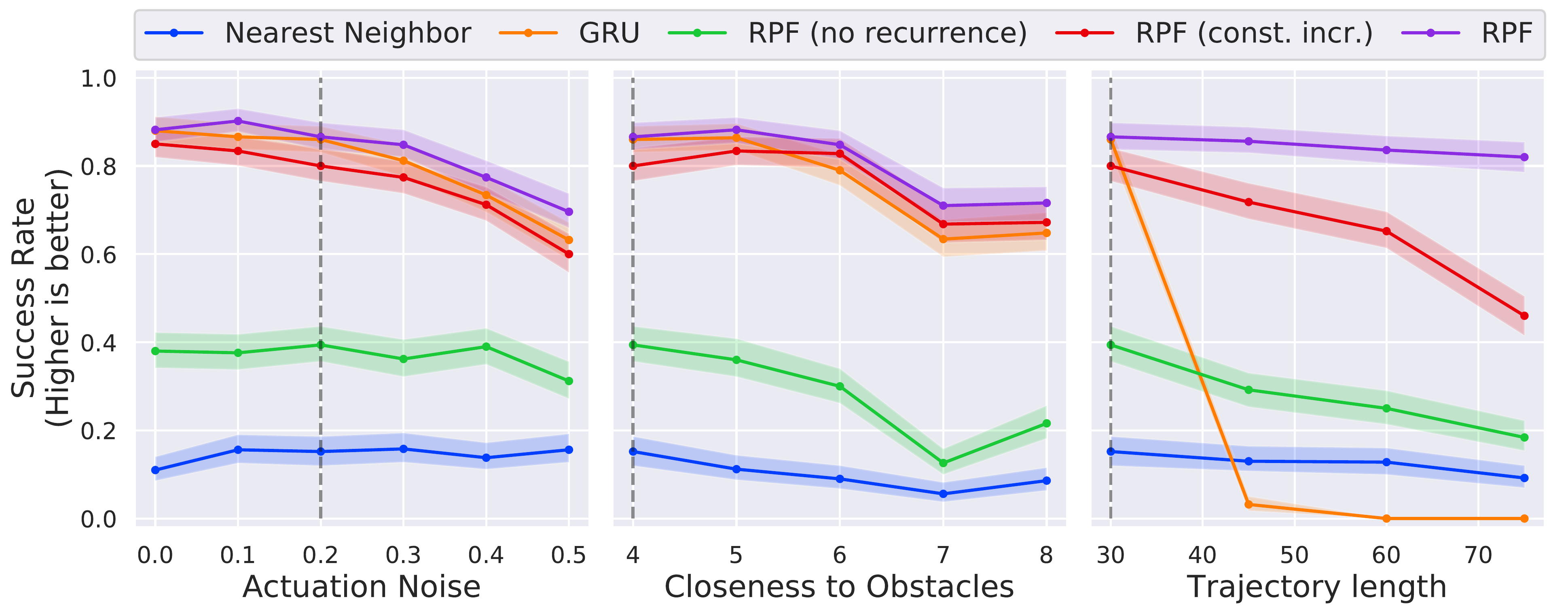} \vspace{-20pt}
    \end{center}
    \caption{{\bf Other Comparisons:} We report success rate for the path
    following task on \sbpd + \mpd for other learning based methods. See text
    for details.}
    \figlabel{other}
\end{wrapfigure}
function. \rpf does better than these other learning methods, and our choice of
modeling $\pi$ as a recurrent function, and learning how to increment the
pointer are important. We also investigated other choices for implementing the
increment function $\sigma$. Increment functions with a small spread around $1$ worked
better than other choices.

\textbf{Discussion.} In this paper, we studied the task of following paths
under noisy actuation and changing environments. 
We operationalized insights from classical robotics with learning-based
models and designed neural architectures that implicitly tackle localization to
output actions to convey the robot directly to the target location.
We made thorough comparisons with geometry-based classical methods and
reasonable learning baselines, and demonstrated the effectiveness of our
proposed approach.

{
\small
\bibliographystyle{plain}
\bibliography{refs-vision,refs-da,refs-robo,refs-nav}

\begin{thebibliography}{10}

\bibitem{anderson2018evaluation}
Peter Anderson, Angel Chang, Devendra~Singh Chaplot, Alexey Dosovitskiy,
  Saurabh Gupta, Vladlen Koltun, Jana Kosecka, Jitendra Malik, Roozbeh
  Mottaghi, Manolis Savva, et~al.
\newblock On evaluation of embodied navigation agents.
\newblock {\em arXiv preprint arXiv:1807.06757}, 2018.

\bibitem{armeni20163d}
Iro Armeni, Ozan Sener, Amir~R Zamir, Helen Jiang, Ioannis Brilakis, Martin
  Fischer, and Silvio Savarese.
\newblock {3D} semantic parsing of large-scale indoor spaces.
\newblock In {\em CVPR}, 2016.

\bibitem{axelrod2017provably}
Brian Axelrod, Leslie~Pack Kaelbling, and Tom{\'a}s Lozano-P{\'e}rez.
\newblock Provably safe robot navigation with obstacle uncertainty.
\newblock In {\em RSS}, 2017.

\bibitem{bowman2017probabilistic}
Sean~L Bowman, Nikolay Atanasov, Kostas Daniilidis, and George~J Pappas.
\newblock Probabilistic data association for semantic slam.
\newblock In {\em ICRA}, 2017.

\bibitem{brahmbhatt2017mapnet}
Samarth Brahmbhatt, Jinwei Gu, Kihwan Kim, James Hays, and Jan Kautz.
\newblock {MapNet}: Geometry-aware learning of maps for camera localization.
\newblock In {\em CVPR}, 2018.

\bibitem{canny1988complexity}
John Canny.
\newblock {\em The complexity of robot motion planning}.
\newblock MIT press, 1988.

\bibitem{Matterport3D}
Angel Chang, Angela Dai, Thomas Funkhouser, Maciej Halber, Matthias Niessner,
  Manolis Savva, Shuran Song, Andy Zeng, and Yinda Zhang.
\newblock {Matterport3D}: Learning from {RGB-D} data in indoor environments.
\newblock In {\em 3DV}, 2017.

\bibitem{chaplot2018active}
Devendra~Singh Chaplot, Emilio Parisotto, and Ruslan Salakhutdinov.
\newblock Active neural localization.
\newblock In {\em ICLR}, 2018.

\bibitem{clement2017robust}
Lee Clement, Jonathan Kelly, and Timothy~D Barfoot.
\newblock Robust monocular visual teach and repeat aided by local ground
  planarity and color-constant imagery.
\newblock {\em JFR}, 2017.

\bibitem{cummins2008fab}
Mark Cummins and Paul Newman.
\newblock {FAB-MAP}: Probabilistic localization and mapping in the space of
  appearance.
\newblock {\em IJRR}, 2008.

\bibitem{daftry2016learning}
Shreyansh Daftry, J~Andrew Bagnell, and Martial Hebert.
\newblock Learning transferable policies for monocular reactive mav control.
\newblock In {\em ISER}, 2016.

\bibitem{davison1998mobile}
Andrew~J Davison and David~W Murray.
\newblock Mobile robot localisation using active vision.
\newblock In {\em ECCV}, 1998.

\bibitem{engel2014lsd}
Jakob Engel, Thomas Sch{\"o}ps, and Daniel Cremers.
\newblock {LSD-SLAM}: Large-scale direct monocular {SLAM}.
\newblock In {\em ECCV}, 2014.

\bibitem{slam-survey:2015}
J.~Fuentes-Pacheco, J.~Ruiz-Ascencio, and J.~M. Rend\'{o}n-Mancha.
\newblock Visual simultaneous localization and mapping: a survey.
\newblock {\em Artificial Intelligence Review}, 2015.

\bibitem{furgale2010visual}
Paul Furgale and Timothy~D Barfoot.
\newblock Visual teach and repeat for long-range rover autonomy.
\newblock {\em JFR}, 2010.

\bibitem{gandhi2017learning}
Dhiraj Gandhi, Lerrel Pinto, and Abhinav Gupta.
\newblock Learning to fly by crashing.
\newblock In {\em IROS}, 2017.

\bibitem{giusti2016machine}
Alessandro Giusti, J{\'e}r{\^o}me Guzzi, Dan~C Cire{\c{s}}an, Fang-Lin He,
  Juan~P Rodr{\'\i}guez, Flavio Fontana, Matthias Faessler, Christian Forster,
  J{\"u}rgen Schmidhuber, Gianni Di~Caro, et~al.
\newblock A machine learning approach to visual perception of forest trails for
  mobile robots.
\newblock {\em RAL}, 2016.

\bibitem{gupta2017cognitive}
Saurabh Gupta, James Davidson, Sergey Levine, Rahul Sukthankar, and Jitendra
  Malik.
\newblock Cognitive mapping and planning for visual navigation.
\newblock In {\em CVPR}, 2017.

\bibitem{izadiUIST11}
Shahram Izadi, David Kim, Otmar Hilliges, David Molyneaux, Richard Newcombe,
  Pushmeet Kohli, Jamie Shotton, Steve Hodges, Dustin Freeman, Andrew Davison,
  and Andrew Fitzgibbon.
\newblock {KinectFusion}: real-time {3D} reconstruction and interaction using a
  moving depth camera.
\newblock {\em UIST}, 2011.

\bibitem{kahn2017self}
Gregory Kahn, Adam Villaflor, Bosen Ding, Pieter Abbeel, and Sergey Levine.
\newblock Self-supervised deep reinforcement learning with generalized
  computation graphs for robot navigation.
\newblock {\em arXiv preprint arXiv:1709.10489}, 2017.

\bibitem{kar2017learning}
Abhishek Kar, Christian H{\"a}ne, and Jitendra Malik.
\newblock Learning a multi-view stereo machine.
\newblock In {\em NIPS}, 2017.

\bibitem{kavraki1996probabilistic}
Lydia~E Kavraki, Petr Svestka, J-C Latombe, and Mark~H Overmars.
\newblock Probabilistic roadmaps for path planning in high-dimensional
  configuration spaces.
\newblock {\em RA}, 1996.

\bibitem{kendall2015posenet}
Alex Kendall, Matthew Grimes, and Roberto Cipolla.
\newblock Posenet: A convolutional network for real-time 6-dof camera
  relocalization.
\newblock In {\em ICCV}, 2015.

\bibitem{khan2017memory}
Arbaaz Khan, Clark Zhang, Nikolay Atanasov, Konstantinos Karydis, Vijay Kumar,
  and Daniel~D Lee.
\newblock Memory augmented control networks.
\newblock In {\em ICLR}, 2018.

\bibitem{lavalle2000rapidly}
Steven~M Lavalle and James~J Kuffner~Jr.
\newblock Rapidly-exploring random trees: Progress and prospects.
\newblock In {\em Algorithmic and Computational Robotics: New Directions},
  2000.

\bibitem{lowe2004distinctive}
David~G Lowe.
\newblock Distinctive image features from scale-invariant keypoints.
\newblock {\em IJCV}, 2004.

\bibitem{mirowski2018learning}
Piotr Mirowski, Matthew~Koichi Grimes, Mateusz Malinowski, Karl~Moritz Hermann,
  Keith Anderson, Denis Teplyashin, Karen Simonyan, Koray Kavukcuoglu, Andrew
  Zisserman, and Raia Hadsell.
\newblock Learning to navigate in cities without a map.
\newblock {\em arXiv preprint arXiv:1804.00168}, 2018.

\bibitem{mirowski2016learning}
Piotr Mirowski, Razvan Pascanu, Fabio Viola, Hubert Soyer, Andy Ballard, Andrea
  Banino, Misha Denil, Ross Goroshin, Laurent Sifre, Koray Kavukcuoglu, et~al.
\newblock Learning to navigate in complex environments.
\newblock In {\em ICLR}, 2017.

\bibitem{nister2004visual}
David Nist{\'e}r, Oleg Naroditsky, and James Bergen.
\newblock Visual odometry.
\newblock In {\em CVPR}, 2004.

\bibitem{parisotto2017neural}
Emilio Parisotto and Ruslan Salakhutdinov.
\newblock {Neural Map}: Structured memory for deep reinforcement learning.
\newblock In {\em ICLR}, 2018.

\bibitem{pathak2018zeroshot}
Deepak Pathak, Parsa Mahmoudieh, Guanghao Luo, Pulkit Agrawal, Dian Chen, Yide
  Shentu, Evan Shelhamer, Jitendra Malik, Alexei~A. Efros, and Trevor Darrell.
\newblock Zero-shot visual imitation.
\newblock In {\em ICLR}, 2018.

\bibitem{pillai2017towards}
Sudeep Pillai and John~J Leonard.
\newblock Towards visual ego-motion learning in robots.
\newblock {\em arXiv preprint arXiv:1705.10279}, 2017.

\bibitem{pomerleau1989alvinn}
Dean~A Pomerleau.
\newblock Alvinn: An autonomous land vehicle in a neural network.
\newblock In {\em NIPS}, 1989.

\bibitem{sadeghi2016cadrl}
Fereshteh Sadeghi and Sergey Levine.
\newblock {(CAD)$^2$RL}: Real singel-image flight without a singel real image.
\newblock In {\em RSS}, 2017.

\bibitem{savinov2018semi}
Nikolay Savinov, Alexey Dosovitskiy, and Vladlen Koltun.
\newblock Semi-parametric topological memory for navigation.
\newblock In {\em ICLR}, 2018.

\bibitem{schonberger2017semantic}
Johannes~L Sch{\"o}nberger, Marc Pollefeys, Andreas Geiger, and Torsten
  Sattler.
\newblock Semantic visual localization.
\newblock In {\em CVPR}, 2018.

\bibitem{schoenberger2016sfm}
Johannes~Lutz Sch\"{o}nberger and Jan-Michael Frahm.
\newblock Structure-from-motion revisited.
\newblock In {\em CVPR}, 2016.

\bibitem{schoenberger2016mvs}
Johannes~Lutz Sch\"{o}nberger, Enliang Zheng, Marc Pollefeys, and Jan-Michael
  Frahm.
\newblock Pixelwise view selection for unstructured multi-view stereo.
\newblock In {\em ECCV}, 2016.

\bibitem{song2016ssc}
Shuran Song, Fisher Yu, Andy Zeng, Angel~X Chang, Manolis Savva, and Thomas
  Funkhouser.
\newblock Semantic scene completion from a single depth image.
\newblock In {\em CVPR}, 2018.

\bibitem{swedish2018deep}
Tristan Swedish and Ramesh Raskar.
\newblock Deep visual teach and repeat on path networks.
\newblock In {\em CVPR}, 2018.

\bibitem{zhang2017neural}
Jingwei Zhang, Lei Tai, Joschka Boedecker, Wolfram Burgard, and Ming Liu.
\newblock Neural slam.
\newblock {\em arXiv preprint arXiv:1706.09520}, 2017.

\bibitem{zhou2017unsupervised}
Tinghui Zhou, Matthew Brown, Noah Snavely, and David~G Lowe.
\newblock Unsupervised learning of depth and ego-motion from video.
\newblock In {\em CVPR}, 2017.

\bibitem{zhu2016target}
Yuke Zhu, Roozbeh Mottaghi, Eric Kolve, Joseph~J Lim, Abhinav Gupta,
  Li~Fei-Fei, and Ali Farhadi.
\newblock Target-driven visual navigation in indoor scenes using deep
  reinforcement learning.
\newblock In {\em ICRA}, 2017.

\end{thebibliography}
}

\end{document}